\documentclass[10pt,journal,compsoc]{IEEEtran}
\usepackage{epsfig}
\usepackage{graphicx}
\usepackage{algorithm}
\usepackage{algorithmic}
\usepackage{amsmath}
\usepackage{amssymb}
\usepackage{mathrsfs} % For LaTeX2e
\usepackage{amsmath,bm}
\usepackage{mdwlist}
\usepackage{array,multirow}
\usepackage{paralist}
\usepackage{bbding}
\usepackage{enumerate}
\usepackage{booktabs}
\usepackage{url}
\usepackage{color}
\usepackage{caption}
\usepackage{xspace}
\usepackage{subfigure}
\usepackage[colorlinks=true,linkcolor=blue,anchorcolor=blue,citecolor=blue]{hyperref}

\def\comment#1{{}}
\def\eg{{\em e.g.}}
\def\ie{{\em i.e.}}
\def\etal{{\em et al.}}

\ifCLASSOPTIONcompsoc
  \usepackage[nocompress]{cite}
\else
  \usepackage{cite}
\fi
\ifCLASSINFOpdf
\else
\fi
\begin{document}
\title{Vision Meets Drones: A Challenge}

\author{Pengfei~Zhu,~Longyin~Wen,~Xiao~Bian,~Haibin~Ling
        and~Qinghua~Hu% <-this % stops a space
\IEEEcompsocitemizethanks{\IEEEcompsocthanksitem Pengfei Zhu and Qinghua Hu are with the School of Computer Science and Technology, Tianjin University, Tianjin, China.
%\protect\\
%E-mail: see http://www.michaelshell.org/contact.html
\IEEEcompsocthanksitem Longyin Wen and Xiao Bian are with GE Global Research, Niskayuna, NY.
\IEEEcompsocthanksitem Haibin Ling is with the Department of Computer \& Information Sciences, Temple University, Philadelphia, PA.
}% <-this % stops an unwanted space
\thanks{}}


\newcommand{\HL}[1]{\textcolor[rgb]{1.00,0.00,.00}{#1}}

\markboth{}%
{Shell \MakeLowercase{\textit{et al.}}: Bare Demo of IEEEtran.cls for Computer Society Journals}


\def \VIS {VisDrone2018}

\IEEEtitleabstractindextext{%
\begin{abstract}
In this paper we present a large-scale visual object detection and tracking benchmark, named {\bf \VIS}, aiming at advancing visual understanding tasks on the drone platform. The images and video sequences in the benchmark were captured over various urban/suburban areas of $14$ different cities across China from north to south. Specifically, {\bf \VIS} consists of $263$ video clips and $10,209$ images (no overlap with video clips) with rich annotations, including object bounding boxes, object categories, occlusion, truncation ratios, etc. With intensive amount of effort, our benchmark has more than $2.5$ million annotated instances in $179,264$ images/video frames. Being the largest such dataset ever published, the benchmark enables extensive evaluation and investigation of visual analysis algorithms on the drone platform. In particular, we design four popular tasks with the benchmark, including object detection in images, object detection in videos, single object tracking, and multi-object tracking. All these tasks are extremely challenging in the proposed dataset due to factors such as occlusion, large scale and pose variation, and fast motion. We hope the benchmark largely boost the research and development in visual analysis on drone platforms.
%\HL{We present a detailed statistical analysis of the dataset in comparison to several other benchmarks in computer vision field, such as MS COCO, KITTI, MOTChallenge, etc. (Not sure if this should be included in the abstract)}

\end{abstract}

}

\maketitle

\IEEEdisplaynontitleabstractindextext

% \fi
\IEEEpeerreviewmaketitle

\IEEEraisesectionheading{
\section{Introduction}
\label{sec:introduction}}

\IEEEPARstart{C}{omputer} vision has been attracting increasing amounts of attention in recent years due to its wide range of applications and recent breakthroughs in many important problems. As two core problems in computer vision, object detection and object tracking are under extensive investigation in both academia and real world applications, \eg, transportation surveillance, smart city, and human-computer interaction. Among many factors and efforts that lead to the fast evolution of computer vision techniques, a notable contribution should be attributed to the invention or organization of numerous benchmarks, such as Caltech \cite{DBLP:journals/pami/DollarWSP12}, KITTI \cite{DBLP:conf/cvpr/GeigerLU12}, ImageNet \cite{DBLP:journals/ijcv/RussakovskyDSKS15}, and MS COCO \cite{DBLP:conf/eccv/LinMBHPRDZ14} for object detection, and OTB \cite{DBLP:journals/pami/WuLY15}, VOT \cite{DBLP:journals/tip/CehovinLK16}, MOTChallenge \cite{DBLP:journals/corr/Leal-TaixeMRRS15}, and UA-DETRAC \cite{DBLP:journals/corr/WenDCLCQLYL15} for object tracking.

Drones (or UAVs) equipped with cameras have been fast deployed to a wide range of applications, including agricultural, aerial photography, fast delivery, surveillance, \textit{etc}. Consequently, automatic understanding of visual data collected from these platforms become highly demanding, which brings computer vision to drones more and more closely. Despite the great progresses in general computer vision algorithms, such as detection and tracking, these algorithms are not usually optimal for dealing with sequences or images captured by drones, due to various challenges such as view point changes and scales. Consequently, developing and evaluating new vision algorithms for drone generated visual data is a key problem in drone-based applications.
However, as pointed out in \cite{DBLP:conf/eccv/MuellerSG16,DBLP:conf/iccv/HsiehLH17}, studies toward this goal is seriously limited by the lack of publicly available large-scale benchmarks or datasets. Some recent efforts \cite{DBLP:conf/eccv/MuellerSG16,DBLP:conf/eccv/RobicquetSAS16,DBLP:conf/iccv/HsiehLH17} have been devoted to construct datasets with drone platform focusing on object detection or tracking. These datasets are still limited in size and scenarios covered, due to the difficulties in data collection and annotation. Thorough evaluations of existing or newly developed algorithms remain an open problem. Thus, a more general and comprehensive benchmark is desired for further boosting visual analysis research on drone platforms.

Thus motivated, we present a large scale benchmark, named {\bf \VIS}, with carefully annotated ground-truth for various important computer vision tasks, to make vision meets drones. The benchmark dataset consists of $263$ video clips formed by $179,264$ frames and $10,209$ static images, captured by various drone-mounted cameras, diverse in a wide range of aspects including location (taken from $14$ different cities in China), environment (urban and country), objects (pedestrian, vehicles, bicycles, \textit{etc}.), and density (sparse and crowded scenes), \textit{etc}. With thorough annotations of over $2.5$ million object instances, the benchmark focuses on four tasks:
\begin{itemize}
\item\textbf{Task 1: object detection in images.} Given a predefined set of object classes (\eg, cars and pedestrians), the task aims to detect objects of these classes from individual images taken from drones.
\item\textbf{Task 2: object detection in videos.} The task is similar to \textbf{Task 1}, except that objects are detected from videos taken from drones.
\item\textbf{Task 3: single object tracking.} The task aims to estimate the state of a target, indicated in the first frame, across frames in an online manner.
\item\textbf{Task 4: multi-object tracking.} The task aims to recover the object trajectories with (\textbf{Task 4B}) or without (\textbf{Task 4A}) the detection results in each video frame.
\end{itemize}
In this challenge we select ten categories of objects of frequent interests in drone applications, such as pedestrians and cars. Altogether we carefully annotated more than $2.5$ million bounding boxes of object instances from these categories. Moreover, some important attributes including visibility of scenes, object category and occlusion, are provided for better data usage. The detailed comparison of the provided drone datasets with other related benchmark datasets in object detection and tracking are presented in Table \ref{tab:comparison-dataset}.

\begin{table*}
\centering
\caption{Comparison of Current State-of-the-Art Benchmarks and Datasets. Note that, the resolution indicates the maximum resolution of the videos/images included in the dataset. }
\label{tab:comparison-dataset}
  \setlength{\tabcolsep}{3.0pt}
% Table 1
\footnotesize{
\begin{tabular}{c|c|c|c|c|c|c|c}
\hline
{\bf Object detection in images}                                                                          &scenario  &\#images  &categories &avg. \#labels/categories &resolution &occlusion labels &year \\
\hline
UIUC \cite{DBLP:journals/pami/AgarwalAR04}                                                 &life &$1,378$ &$1$ &$739$ &$200\times150$ & &2004 \\
INRIA \cite{DBLP:conf/cvpr/DalalT05}                                                               &life &$2,273$ &$1$ &$1,774$ &$96\times160$ & &2005 \\
ETHZ Pedestrian \cite{DBLP:conf/iccv/EssLG07}                                             &life &$2,293$ &$1$ &$10.9k$ &$640\times480$ & &2007 \\
TUD \cite{DBLP:conf/cvpr/AndrilukaRS08}                                                        &life &$1,818$ &$1$ &$3,274$ &$640\times480$ & &2008 \\
EPFL Multi-View Car \cite{DBLP:conf/cvpr/OzuysalLF09}                                 &exhibition &$2,000$ &$1$ &$2,000$ &$376\times250$ & &2009 \\
Caltech Pedestrian \cite{DBLP:journals/pami/DollarWSP12}                             &driving &$249k$ &$1$ &$347k$ &$640\times480$ &$\surd$ &2012 \\
KITTI Detection \cite{DBLP:conf/cvpr/GeigerLU12}                                            &driving &$15.4k$ &$2$ &$80k$ &$1241\times376$ &$\surd$ &2012 \\
PASCAL VOC2012 \cite{pascal-voc-2012}                                                         &life &$22.5k$ &$20$ &$1,373$ &$469\times387$ &$\surd$ &2012 \\
ImageNet Object Detection \cite{DBLP:journals/ijcv/RussakovskyDSKS15}      &life &$456.2k$ &$200$ &$2,007$ &$482\times415$ &$\surd$ &2013 \\
MS COCO \cite{DBLP:conf/eccv/LinMBHPRDZ14}                                            &life &$328.0k$ &$91$ &$27.5k$ &$640\times640$ & &2014 \\
VEDAI \cite{DBLP:journals/jvcir/RazakarivonyJ16}                                            &satellite &$1.2k$ &$9$ &$733$ &$1024\times1024$ & &2015 \\
COWC \cite{DBLP:conf/eccv/MundhenkKSB16}                                                &aerial   &$32.7k$  &$1$ &$32.7k$ &$2048\times2048$ & &2016\\
CARPK \cite{DBLP:conf/iccv/HsiehLH17}                                                          &drone &$1,448$ &1 &$89.8k$ &$1280\times720$ & &2017 \\
{\bf \VIS}                                                                                              &drone &$10,209$ &$10$ &$54.2k$ &$2000\times1500$
 &$\surd$ &2018 \\
\hline
\end{tabular}}

\begin{tabular}{c}
\\
\\
\end{tabular}

% Table 2
  \setlength{\tabcolsep}{3.5pt}
\footnotesize{
\begin{tabular}{c|c|c|c|c|c|c|c}
\hline
{\bf Object detection in videos} &scenario &\#frames &categories &avg. \#labels/categories &resolution &occlusion labels &year \\
\hline
ImageNet Video Detection \cite{DBLP:journals/ijcv/RussakovskyDSKS15}      &life &$2017.6k$ &$30$ &$66.8k$ &$1280\times1080$ &$\surd$ &2015 \\
UA-DETRAC Detection \cite{DBLP:journals/corr/WenDCLCQLYL15}                &surveillance &$140.1k$ &$4$ &$302.5k$ &$960\times540$ &$\surd$ &2015 \\
MOT17Det \cite{MOT17}      &life &$11.2k$ &$1$ &$392.8k$ &$1920\times1080$ &$\surd$ &2017 \\
Okutama-Action \cite{DBLP:conf/cvpr/BarekatainMSMNM17}      &drone &$77.4k$ &$1$ &$422.1k$ &$3840\times2160$ & &2017 \\
{\bf \VIS}                                                                                             &drone &$40.0k$ &$10$ &$183.3k$ &$3840\times2160$ &$\surd$ &2018 \\
\hline
\end{tabular}}

\begin{tabular}{c}
\\
\\
\end{tabular}

% Table 3
  \setlength{\tabcolsep}{25.0pt}
\footnotesize{
\begin{tabular}{c|c|c|c|c}
\hline
{\bf Single object tracking}                                               &scenarios &\#sequences &\#frames &year \\
\hline

ALOV3000 \cite{DBLP:journals/pami/SmeuldersCCCDS14}                      &life &$314$ &$151.6k$ &2014 \\
OTB100 \cite{DBLP:journals/pami/WuLY15}                      &life &$100$ &$59.0k$ &2015 \\
TC128 \cite{DBLP:journals/tip/LiangBL15}    & life & $128$ &$55.3k$ &2015 \\
VOT2016 \cite{DBLP:conf/eccv/KristanLMFPCVHL16}     &life &$60$ &$21.5k$ &2016 \\
UAV123 \cite{DBLP:conf/eccv/MuellerSG16}                    &drone &$123$ &$110k$ &2016 \\
NfS \cite{DBLP:conf/iccv/GaloogahiFHRL17}                    &life &$100$ &$383k$ &2017 \\
POT 210 \cite{LiangWLWLL18icra}                    &planar objects &$210$ &$105.2k$ &2018 \\
{\bf \VIS}                                                             &drone &$167$ &$139.3k$ &2018 \\
\hline
\end{tabular}}

\begin{tabular}{c}
\\
\\
\end{tabular}

% Table 4
  \setlength{\tabcolsep}{2.0pt}
\footnotesize{
\begin{tabular}{c|c|c|c|c|c|c|c}
\hline
{\bf Multi-object tracking} &scenario &\#frames &categories &avg. \#labels/categories &resolution &occlusion labels &year \\
\hline
KITTI Tracking \cite{DBLP:conf/cvpr/GeigerLU12}                                              &driving &$19.1k$ &$5$ &$19.0k$ &$1392\times512$  &$\surd$ &2013 \\
MOTChallenge 2015 \cite{DBLP:journals/corr/Leal-TaixeMRRS15}                    &surveillance &$11.3k$ &$1$ &$101.3k$ &$1920\times1080$ & &2015 \\
UA-DETRAC Tracking \cite{DBLP:journals/corr/WenDCLCQLYL15}                  &surveillance &$140.1k$ &$4$ &$302.5k$ &$960\times540$ &$\surd$ &2015 \\
DukeMTMC \cite{DBLP:conf/eccv/RistaniSZCT16}                  &surveillance &$2852.2k$ &$1$ &$4077.1k$ &$1920\times1080$ & &2016 \\
Campus \cite{DBLP:conf/eccv/RobicquetSAS16}                                                &drone &$929.5k$ &$6$ &$1769.4k$ &$1417\times2019$ & &2016\\
MOT17 \cite{MOT17}                                             &surveillance &$11.2k$ &$1$ &$392.8k$ &$1920\times1080$ & &2017 \\
{\bf \VIS}                                                                                                &drone &$40.0k$ &$10$ &$183.3k$ &$3840\times2160$ &$\surd$ &2018 \\
\hline
\end{tabular}}
\label{tab:comparison-datasets}
\end{table*}

\section{Related Work}
In recent years, the computer vision community has developed various benchmarks for numerous tasks, including generic object detection \cite{DBLP:conf/eccv/LinMBHPRDZ14,DBLP:journals/ijcv/RussakovskyDSKS15}, pedestrian detection \cite{DBLP:journals/pami/DollarWSP12}, single object tracking \cite{DBLP:journals/pami/WuLY15,DBLP:conf/eccv/KristanLMFPCVHL16}, multi-object tracking \cite{DBLP:journals/corr/Leal-TaixeMRRS15,DBLP:journals/corr/WenDCLCQLYL15}, 3D reconstruction \cite{DBLP:conf/cvpr/SeitzCDSS06}, and optical flow \cite{DBLP:journals/ijcv/BakerSLRBS11,DBLP:conf/cvpr/GeigerLU12}, which are extremely helpful to advance the state of the art in the respective areas. In this section, we review the most relevant drone-based benchmarks and other benchmarks in object detection and tracking fields.

\begin{figure*}[t]
\centering
\includegraphics[width=1.0\linewidth]{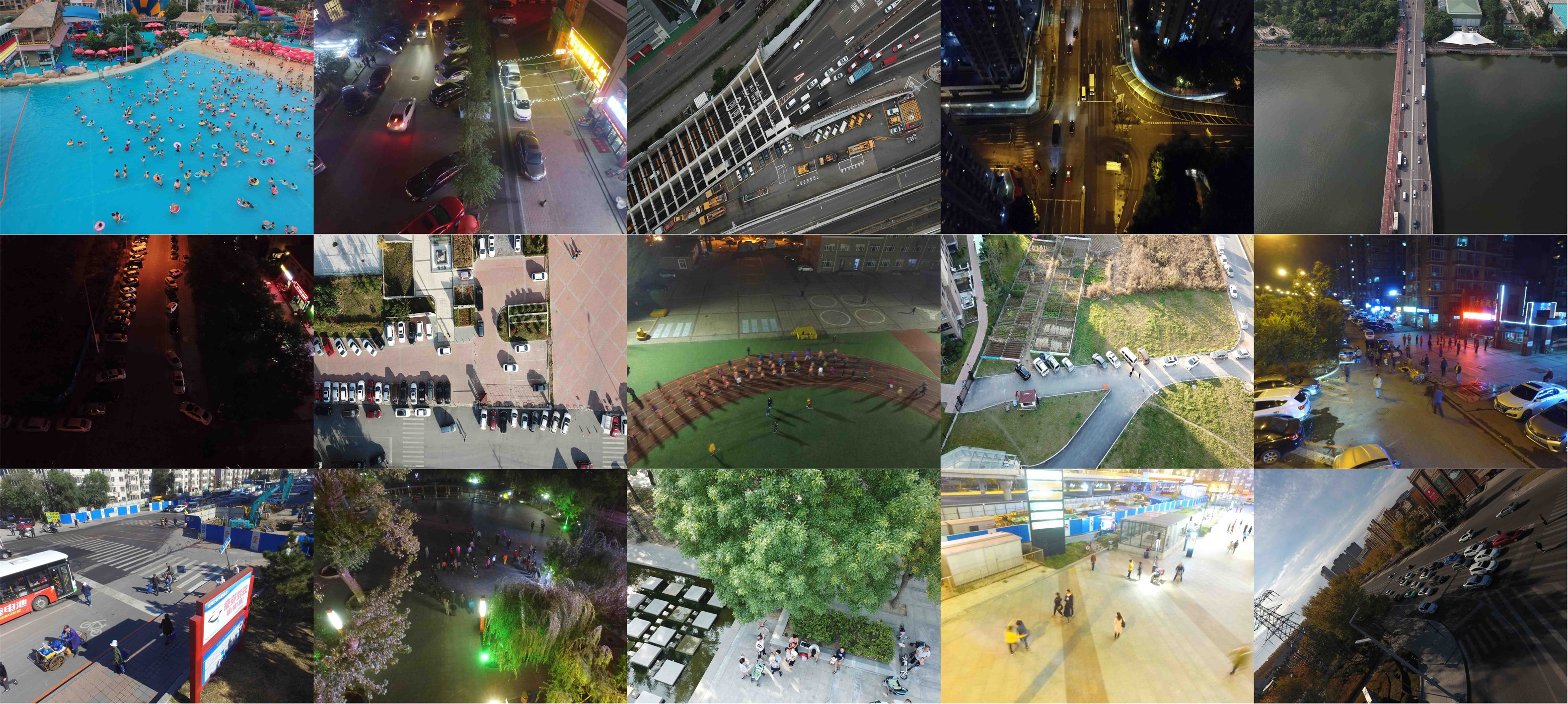}
\caption{Some example static images for {\bf Task 1} (object detection in images) in the {\bf \VIS} challenge.}
\label{fig:image_sample}
\end{figure*}

\subsection{Drone-based Datasets}
To date, there only exists a handful of drone-based datasets in computer vision field. Hsieh \etal \cite{DBLP:conf/iccv/HsiehLH17} present a dataset for car counting, which consists of $1,448$ images captured in parking lot scenarios with the drone platform, including $89,777$ annotated cars. Robicquet \etal \cite{DBLP:conf/eccv/RobicquetSAS16} collect several video sequences with the drone platform in campuses, including various types of objects, (\ie, pedestrians, bikes, skateboarders, cars, buses, and golf carts), which enable the design of new object tracking and trajectory forecasting algorithms. Barekatain \cite{DBLP:conf/cvpr/BarekatainMSMNM17} present a new Okutama-Action dataset for concurrent human action detection with the aerial view. The dataset includes $43$ minute-long fully-annotated sequences with $12$ action classes. In  \cite{DBLP:conf/eccv/MuellerSG16}, a high-resolution UAV123 dataset is presented for single object tracking, which contains $123$ aerial video sequences with $110k$ ($1k=1,000$) fully annotated frames, including the bounding boxes of people and their corresponding action labels. Li \etal \cite{DBLP:conf/aaai/LiY17} capture $70$ video sequences of high diversity by drone cameras and manually annotate the bounding boxes of objects for single object tracking evaluation. In \cite{DBLP:journals/pami/RozantsevLF17}, Rozantsev \etal present two separate datasets for detecting flying objects, \ie, the UAV dataset and the aircraft dataset. The former one comprises $20$ video sequences with the resolution $752\times480$ and $8,000$ annotated bounding boxes of objects, acquired by a camera mounted on a drone flying indoors and outdoors. The latter one consists of $20$ publicly available videos of radio-controlled planes with $4,000$ annotated bounding boxes. In contrast to the aforementioned datasets acquired in constrained scenarios for single object tracking or object detection and counting, our {\bf \VIS} dataset is captured in various unconstrained urban scenes, focusing on four core problems in computer vision fields, \ie, object detection in images, object detection in videos, single object tracking, and multi-object tracking.

\subsection{Object Detection Datasets}
Several object detection benchmarks have been collected for evaluating object detection algorithms. Enzweiler and Gavrila \cite{DBLP:journals/pami/EnzweilerG09} present the Daimler dataset, captured by a vehicle driving through urban environment. The dataset includes $3,915$ manually annotated pedestrians in video images in the training set, and $21,790$ video images with $56,492$ annotated pedestrians in the testing set. Caltech \cite{DBLP:journals/pami/DollarWSP12} consists of approximately $10$ hours of $640\times480$ 30Hz videos taken from a vehicle driving through regular traffic in an urban environment. It contains $\sim250,000$ frames with a total of $350,000$ annotated bounding boxes of $2,300$ unique pedestrians. KITTI-D \cite{DBLP:conf/cvpr/GeigerLU12} is designed to evaluate the car, pedestrian, and cyclist detection algorithms in autonomous driving scenarios, with $7,481$ training and $7,518$ testing images. Mundhenk \etal \cite{DBLP:conf/eccv/MundhenkKSB16} create a  large dataset for classification, detection and counting of cars, which contains $32,716$ unique cars from six different image sets, each covering a different geographical location and produced by different imagers. The recent UA-DETRAC benchmark \cite{DBLP:journals/corr/WenDCLCQLYL15,DBLP:conf/avss/LyuCDWQLWKHCCAB17}  provides $1,210k$ objects in $140k$ frames for vehicle detection.

The PASCAL VOC dataset  \cite{DBLP:journals/ijcv/EveringhamGWWZ10,DBLP:journals/ijcv/EveringhamEGWWZ15} is one of the pioneering work in generic object detection filed, which is designed to provide a standardized test bed for object detection, image classification, object segmentation, person layout, and action classification. ImageNet \cite{DBLP:conf/cvpr/DengDSLL009,DBLP:journals/ijcv/RussakovskyDSKS15} follows the footsteps of the PASCAL VOC dataset by scaling up more than an order of magnitude in number of object classes and images, \ie, PASCAL VOC 2012 has $20$ object classes and $21,738$ images {\em vs.} ILSVRC2012 with $1,000$ object classes and $1,431,167$ annotated images. Recently, Lin \etal \cite{DBLP:conf/eccv/LinMBHPRDZ14} release the MS COCO dataset, containing more than $328,000$ images with $2.5$ million manually segmented object instances. It has $91$ object categories with $27k$ instances on average per category. Notably, it contains object segmentation annotations which are not available in ImageNet.

\begin{figure*}[t]
\centering
\includegraphics[width=1.0\linewidth]{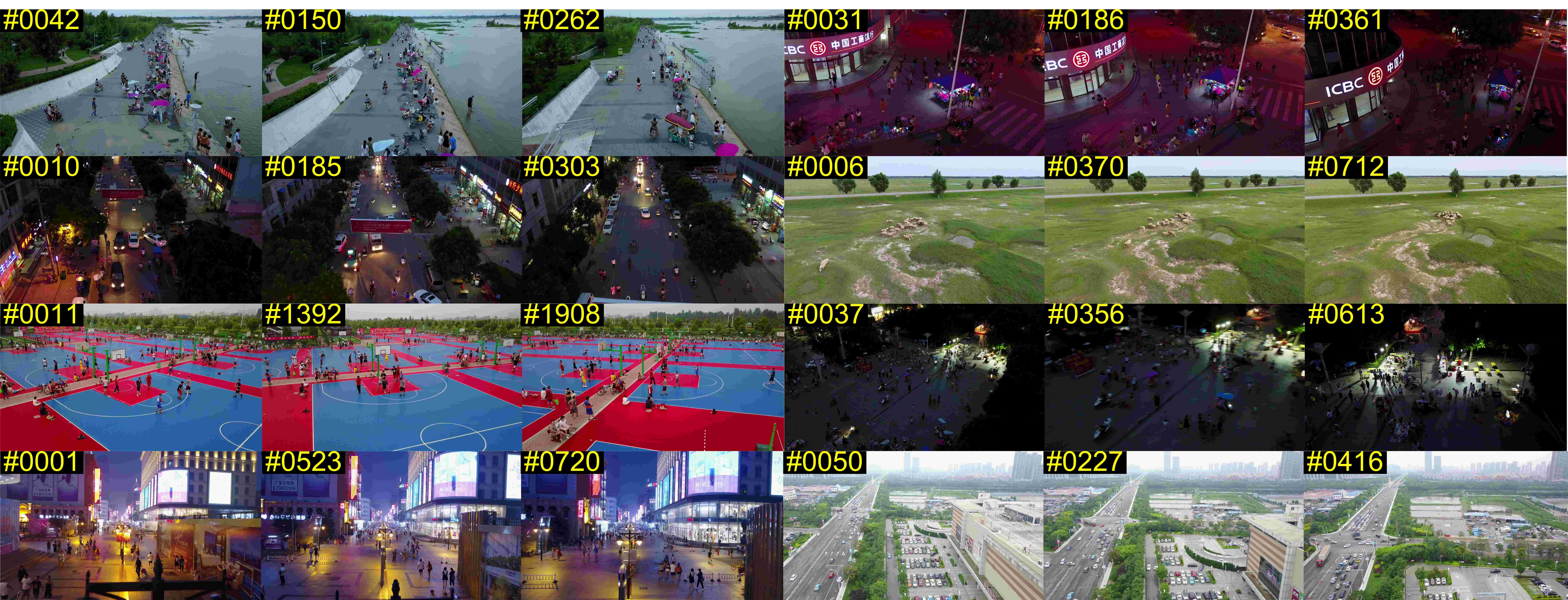}
\caption{Some example screenshots of video clips for {\bf Task 2} (object detection in videos), {\bf Task 3} (single object tracking), and {\bf Task 4} (multi-object tracking) in the {\bf \VIS} challenge. The frame index is placed on the left top corner of each screenshot.}
\label{fig:video_sample}
\end{figure*}

\subsection{Object Tracking Datasets}

{\flushleft \textbf{Single-object tracking.}}
Single object tracking is one of the fundamental problems in computer vision, which aims to estimate the trajectory of a target in a video sequence, with its given initial state. In recent years, numerous datasets have been developed for single object tracking evaluation. Wu \etal \cite{DBLP:conf/cvpr/WuLY13} develop a standard platform to evaluate the single object tracking algorithms, and scale up the data size from $50$ sequences to $100$ sequences in \cite{DBLP:journals/pami/WuLY15}. Similarly, Liang \etal \cite{DBLP:journals/tip/LiangBL15} collect $128$ video sequences for evaluating the color enhanced trackers. To track the progress in visual tracking field, Kristan \etal \cite{DBLP:conf/eccv/KristanPLMCNVFL14,DBLP:conf/iccvw/KristanMLFCFVHN15,DBLP:conf/eccv/KristanLMFPCVHL16} organize a VOT competition from $2013$ to $2017$ by presenting new datasets and evaluation strategies for tracking evaluation. Smeulders \etal \cite{DBLP:journals/pami/SmeuldersCCCDS14} present the ALOV300 dataset, which contains $314$ video sequences with $14$ visual attributes, such as long duration, zooming camera, moving camera and transparency. Li \etal \cite{DBLP:journal/pami/LiAN15} construct a large-scale dataset with $365$ video sequences of pedestrians and rigid objects, covering $12$ kinds of objects captured from moving cameras. Du \etal \cite{DBLP:journals/tip/DuQLWHL16} design a dataset including $50$ annotated video sequences, focusing on deformable object tracking in unconstrained environments. To evaluate tracking algorithms in higher frame rate video sequences, Galoogahi \etal \cite{DBLP:conf/iccv/GaloogahiFHRL17} propose a dataset including $100$ videos ($380k$ frames) recorded by the higher frame rate cameras ($240$ frame per second) from real world scenarios. Besides using video sequences captured by RGB cameras,  Felsberg \etal \cite{DBLP:conf/iccvw/FelsbergBHAKMLC15,DBLP:conf/eccv/FelsbergKMLPHBE16} organize a series of competitions from 2015 to 2017, focusing on visual tracking on thermal video sequences recorded by eight different types of sensors. In \cite{DBLP:conf/iccv/SongX13}, a RGB-D tracking dataset is presented, which includes $100$ video clips with RGB and depth channels and manually annotated ground truth bounding boxes.

{\flushleft \textbf{Multi-object tracking.}}
Multi-object tracking is another important research problem with many applications, such as surveillance, behavior analysis, and autonomous driving. Some of the most widely used multi-object tracking evaluation datasets include the PETS09 \cite{DBLP:conf/avss/FerrymanE09}, PETS16 \cite{website:pets2016/pets2016}, KITTI-T \cite{DBLP:conf/cvpr/GeigerLU12}, MOTChallenge \cite{DBLP:journals/corr/Leal-TaixeMRRS15,DBLP:journals/corr/Anton16}, and UA-DETRAC \cite{DBLP:journals/corr/WenDCLCQLYL15,DBLP:conf/avss/LyuCDWQLWKHCCAB17}. Specifically, the PETS09 \cite{DBLP:conf/avss/FerrymanE09} and PETS16 \cite{website:pets2016/pets2016} datasets mainly focus on multi-pedestrian detection, tracking and counting in the surveillance scenarios. KITTI-T \cite{DBLP:conf/cvpr/GeigerLU12} is designed for object tracking in autonomous driving, which is recorded from a moving vehicle with viewpoint of the driver. MOT15 \cite{DBLP:journals/corr/Leal-TaixeMRRS15} and MOT16 \cite{DBLP:journals/corr/Anton16} aim to provide a unified dataset, platform, and evaluation protocol for multiple object tracking algorithms, including $22$ and $14$ sequences respectively. Recently, the UA-DETRAC benchmark \cite{DBLP:journals/corr/WenDCLCQLYL15,DBLP:conf/avss/LyuCDWQLWKHCCAB17} is constructed, which contains a total of $100$ sequences to track multiple vehicles, where sequences are filmed from a surveillance viewpoint.

Moreover, in some scenarios, a network of cameras are set up to capture multi-view information to help multi-object tracking. The datasets in \cite{DBLP:journals/pami/FleuretBLF08,DBLP:conf/avss/FerrymanE09} are recorded using multi-camera with fully overlapping views in constrained environments. Other datasets are captured by non-overlapping cameras. For example, Chen \etal \cite{DBLP:journals/tcsv/ChenCCH17} collect four datasets, each of which includes $3$ to $5$ cameras with non-overlapping views in real scenes and simulation environments. In \cite{DBLP:conf/eccv/KuoHN10}, the dataset is captured by $3$ cameras in the campus environments with the resolution of $852\times480$ and $25$ minutes length. Zhang \etal \cite{DBLP:conf/wacv/ZhangSFR15} develop a dataset composed of $5$ to $8$ cameras covering both indoor and outdoor scenes at a university. Ristani \etal \cite{DBLP:conf/eccv/RistaniSZCT16} organize a challenge and present a large-scale fully-annotated and calibrated dataset, including more than $2$ million 1080p video frames taken by $8$ cameras with more than $2,700$ identities.

\begin{figure*}[t]
\centering
\includegraphics[width=0.9\linewidth]{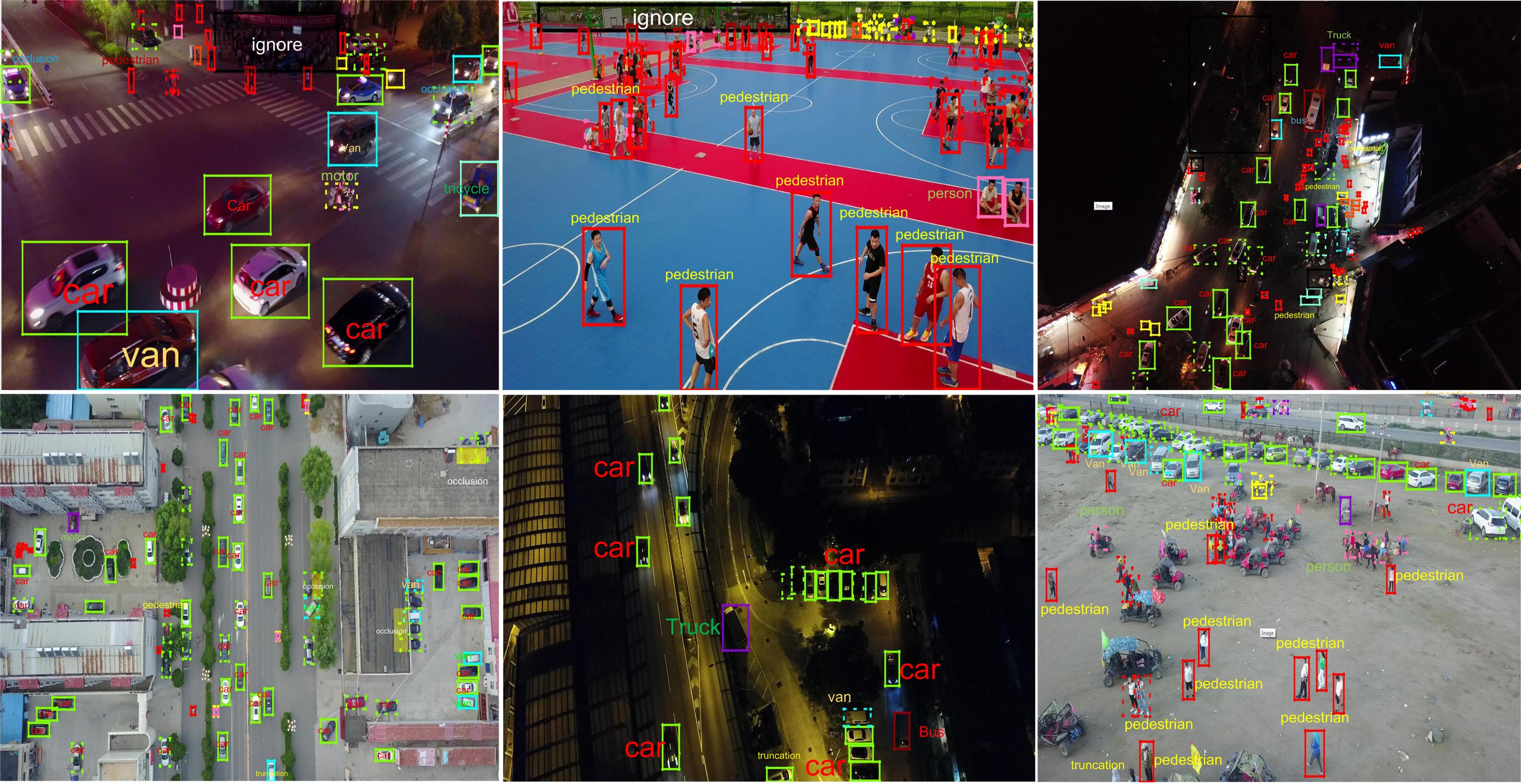}
\caption{Some annotated example images of (\textbf{Task 1}) object detection in images. The dashed bounding box indicates the object is occluded. Different bounding box colors indicate different classes of objects. For better visualization, we only display some attributes. }
\label{fig:t1_annotation}
\end{figure*}

\section{VisDrone2018 Benchmark Dataset}

\subsection{Dataset Collection}
A critical basis for effective algorithm evaluation is a thorough dataset. For this purpose, in {\bf \VIS}, we systematically collected the largest, to the best of our knowledge, drone image/video dataset. Our dataset consists of $263$ video clips with $179,264$ frames and additional $10,209$ static images. The videos/images are acquired by various drone platforms, \ie, DJI Mavic, Phantom series (3, 3A, 3SE, 3P, 4, 4A, 4P), including different scenarios across $14$ different cites in China, \ie, Tianjin, Hongkong, Daqing, Ganzhou, Guangzhou, Jincang, Liuzhou, Nanjing, Shaoxing, Shenyang, Nanyang, Zhangjiakou, Suzhou and Xuzhou. The dataset covers various weather and lighting conditions, representing diverse scenarios in our daily life. The maximal resolutions of video clips and static images are $3840\times2160$ and $2000\times1500$, respectively. Some example images and video clips are shown in Figures \ref{fig:image_sample} and \ref{fig:video_sample}.

A website: \url{www.aiskyeye.com} is constructed for accessing the {\bf \VIS} benchmark and perform evaluation of the four tasks, \ie, (1) object detection in images, (2) object detection in videos, (3) single object tracking, and (4) multi-object tracking. Specifically, a user is required to create an account using an institutional email address. After registration, the user can choose the tasks which she or he decides to participate, and submit the results using the corresponding account. Notably, for each task, the images/videos in the training, validation, and testing sets are captured at different locations, but with similar scenarios. The manually annotated ground truths for training and validation are made available to users, but the ground truths of the testing set are reserved in order to avoid (over)fitting of algorithms. We encourage the participants to use the provided training data, but also allow them to use additional training data. The use of additional training data must be indicated during submission. In the following subsections, we describe each task and the corresponding data annotation and statistics in details.

\begin{figure*}[t]
\centering
\includegraphics[width=0.9\linewidth]{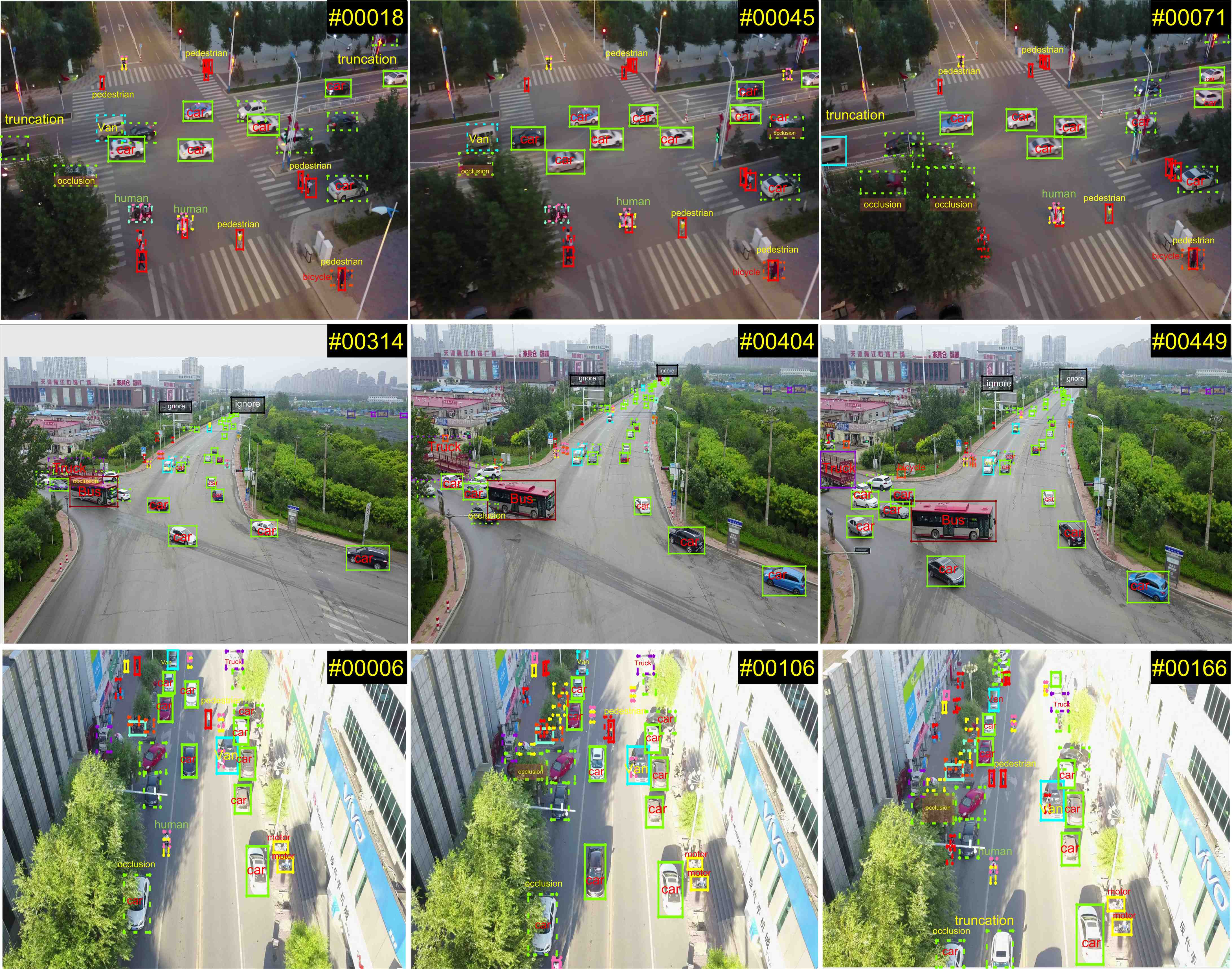}
%\vspace{-2mm}
\caption{Some annotated example video frames of (\textbf{Task 2}) object detection in videos. The dashed bounding box indicates the object is occluded. Different bounding box colors indicate different classes of objects. For better visualization, we only display some attributes.}
\label{fig:t2_annotation}
\end{figure*}

\subsection{Task 1: Object Detection in Images}
\begin{figure*}[t]
\centering
\includegraphics[width=0.8\linewidth]{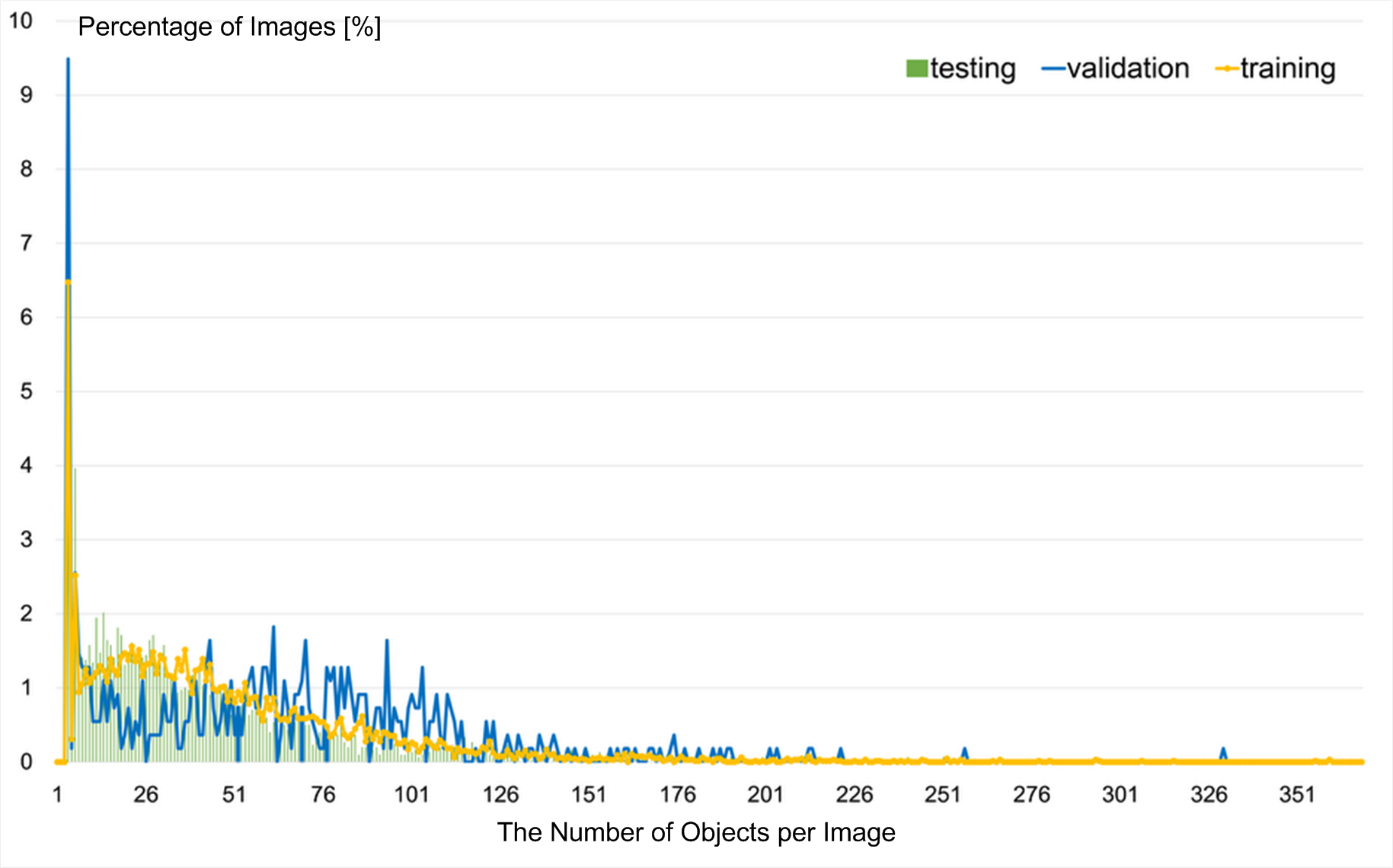}
\vspace{-2mm}
\caption{The number of objects per image {\em vs.} percentage of images in the training, validation and testing sets for ({\bf Task 1}) object detection in images.}
\label{fig:image_number_distribution}
\end{figure*}

Given an input image and a predefined set of object categories, \eg, car and pedestrian, the task of object detection (in images) aims to locate all the object instances in these categories from the image (if any). Typically and in our benchmark, for each object class, we require a detection algorithm to predict the bounding box of each instance of that class in the image, with a real-valued confidence.  The {\bf \VIS} provides a dataset of $10,209$ images for this task, with $6,471$ images used for training, $548$ for validation and $3,190$ for testing. The images of the three subsets are taken at different locations, but share similar environments and attributes. We plot the number of objects per image {\em vs.} percentage of images in each subset to show the distributions of the number of objects in each image of the training, validation and testing sets in Figure \ref{fig:image_number_distribution}.

\begin{figure*}
\centering
\includegraphics[width=1.0\linewidth]{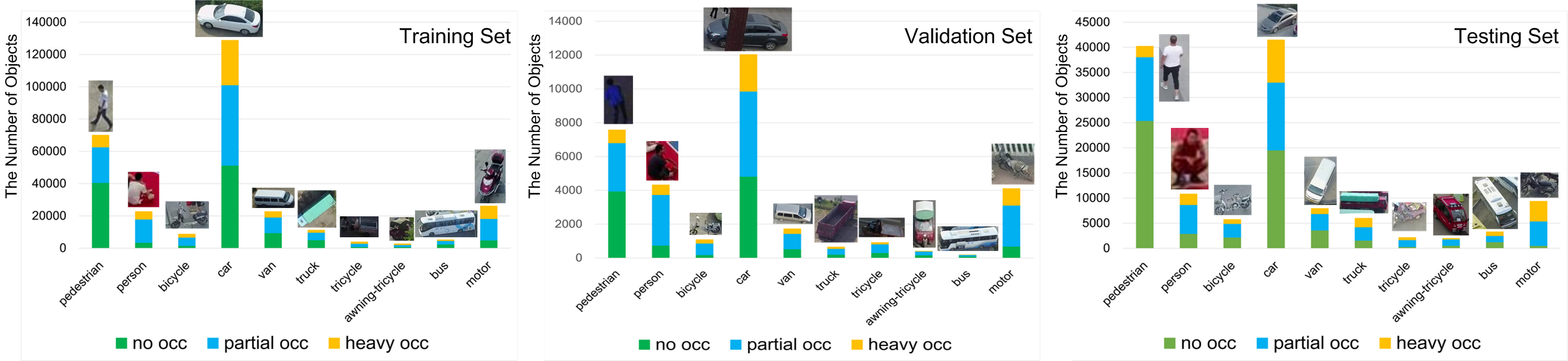}
\caption{The number of objects with different occlusion degrees of different object categories in the training, validation and testing sets for ({\bf Task 1}) object detection in images.}
\label{fig:image_annotation_statistics}
\end{figure*}

For object categories, we mainly focus on human and vehicles in our daily life, and define ten object categories of interest including {\em pedestrian}, {\em person}\footnote{If a human maintains standing pose or walking, we classify it as {\em pedestrian}; otherwise, it is classified as a {\em person}.}, {\em car}, {\em van}, {\em bus}, {\em truck}, {\em motor}, {\em bicycle}, {\em awning-tricycle}, and {\em tricycle}. Some rarely occurring special vehicles (\eg, machineshop truck, forklift truck, and tanker) are ignored in evaluation. We manually annotate the bounding boxes of different categories of objects in each image. In addition, we also provide two kinds of useful annotations, occlusion ratio and truncation ratio. Specifically, we use the fraction of objects being occluded to define the occlusion ratio, and define three degrees of occlusions: no occlusion (occlusion ratio $0\%$), partial occlusion (occlusion ratio $1\%\sim50\%$), and heavy occlusion (occlusion ratio $>50\%$). For truncation ratio, it is used to indicate the degree of object parts appears outside a frame. If an object is not fully captured within a frame, we annotate the bounding box across the frame boundary and estimate the truncation ratio based on the region outside the image. It is worth mentioning that a target is skipped during evaluation if its truncation ratio is larger than $50\%$. We show some annotated examples in Figure \ref{fig:t1_annotation}, and present the number of objects with different occlusion degrees of different object categories in the training, validation, and testing sets in Figure \ref{fig:image_annotation_statistics}.

\begin{figure*}[t]
\centering
\includegraphics[width=0.9\linewidth]{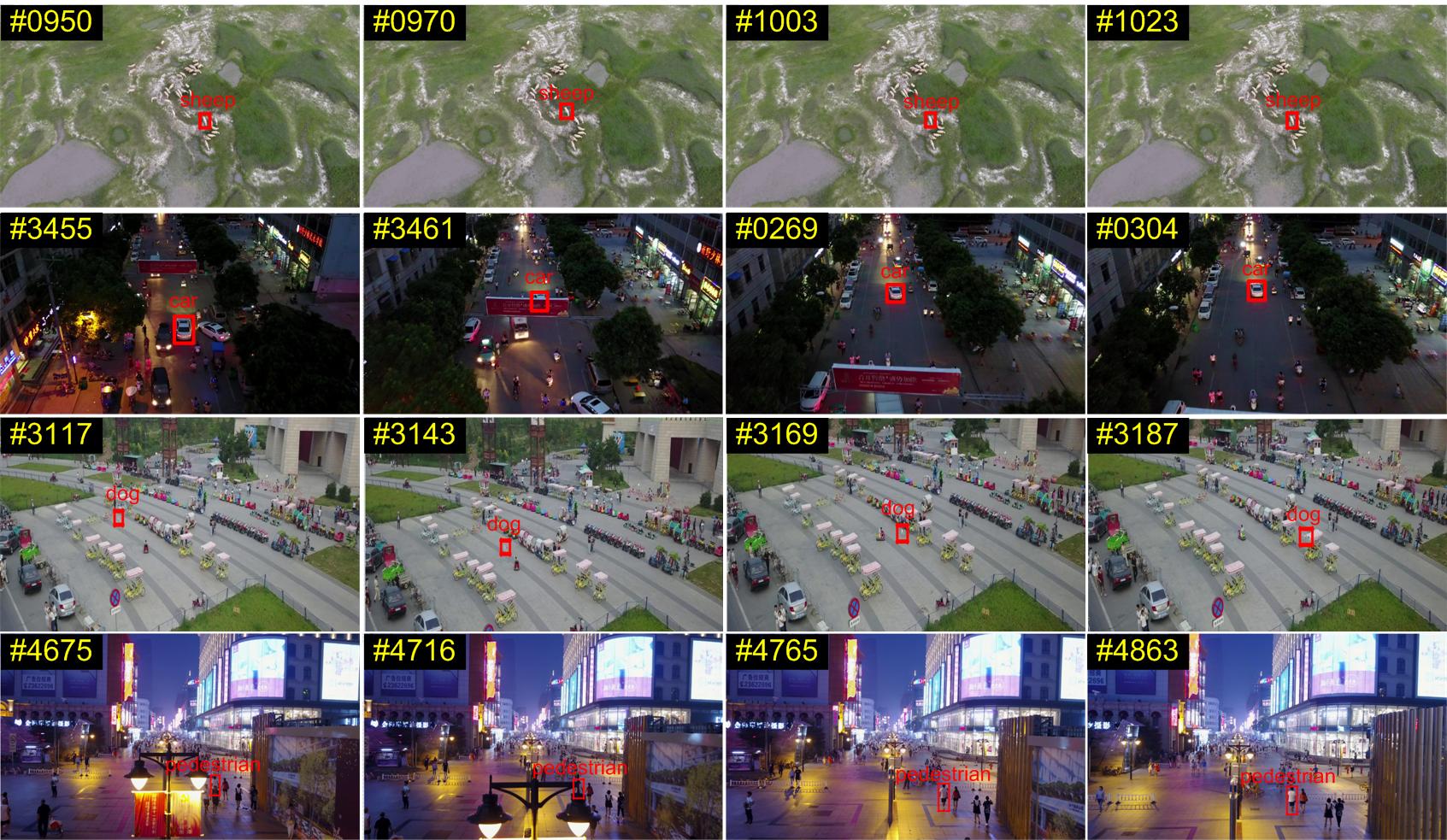}
\caption{Some annotated example video frames of (\textbf{Task 3}) single object tracking. }
\label{fig:t3_annotation}
\end{figure*}

\subsubsection{Evaluation Criteria}
We require each evaluated algorithm in {\bf Task 1} (object detection in images) to output a list of detected bounding boxes with confidence scores for each test image. Following the evaluation protocol in MS COCO \cite{DBLP:conf/eccv/LinMBHPRDZ14}, we use the AP$^{\text{IoU}=0.50:0.05:0.95}$, AP$^{\text{IoU}=0.50}$, AP$^{\text{IoU}=0.75}$, AR$^{\text{max}=1}$, AR$^{\text{max}=10}$, AR$^{\text{max}=100}$ and AR$^{\text{max}=500}$ metrics to evaluate the results of detection algorithms. These criteria penalize missing detection of objects as well as duplicate detections (two detection results for the same object instance). Specifically, AP$^{\text{IoU}=0.50:0.05:0.95}$ is computed by averaging over all $10$ intersection over union (IoU) thresholds (\ie, in the range $[0.50:0.95]$ with the uniform step size $0.05$) of all categories, which is used as the primary metric for ranking. AP$^{\text{IoU}=0.50}$ and AP$^{\text{IoU}=0.75}$ are computed at the single IoU thresholds $0.5$ and $0.75$ over all categories, respectively. The AR$^{\text{max}=1}$, AR$^{\text{max}=10}$, and AR$^{\text{max}=100}$ scores are the maximum recalls given $1$, $10$, $100$ and $500$ detections per image, averaged over all categories and IoU thresholds. Please refer to \cite{DBLP:conf/eccv/LinMBHPRDZ14} for more details.

\subsection{Task 2: Object Detection in Videos}
\begin{figure*}
\centering
\includegraphics[width=0.8\linewidth]{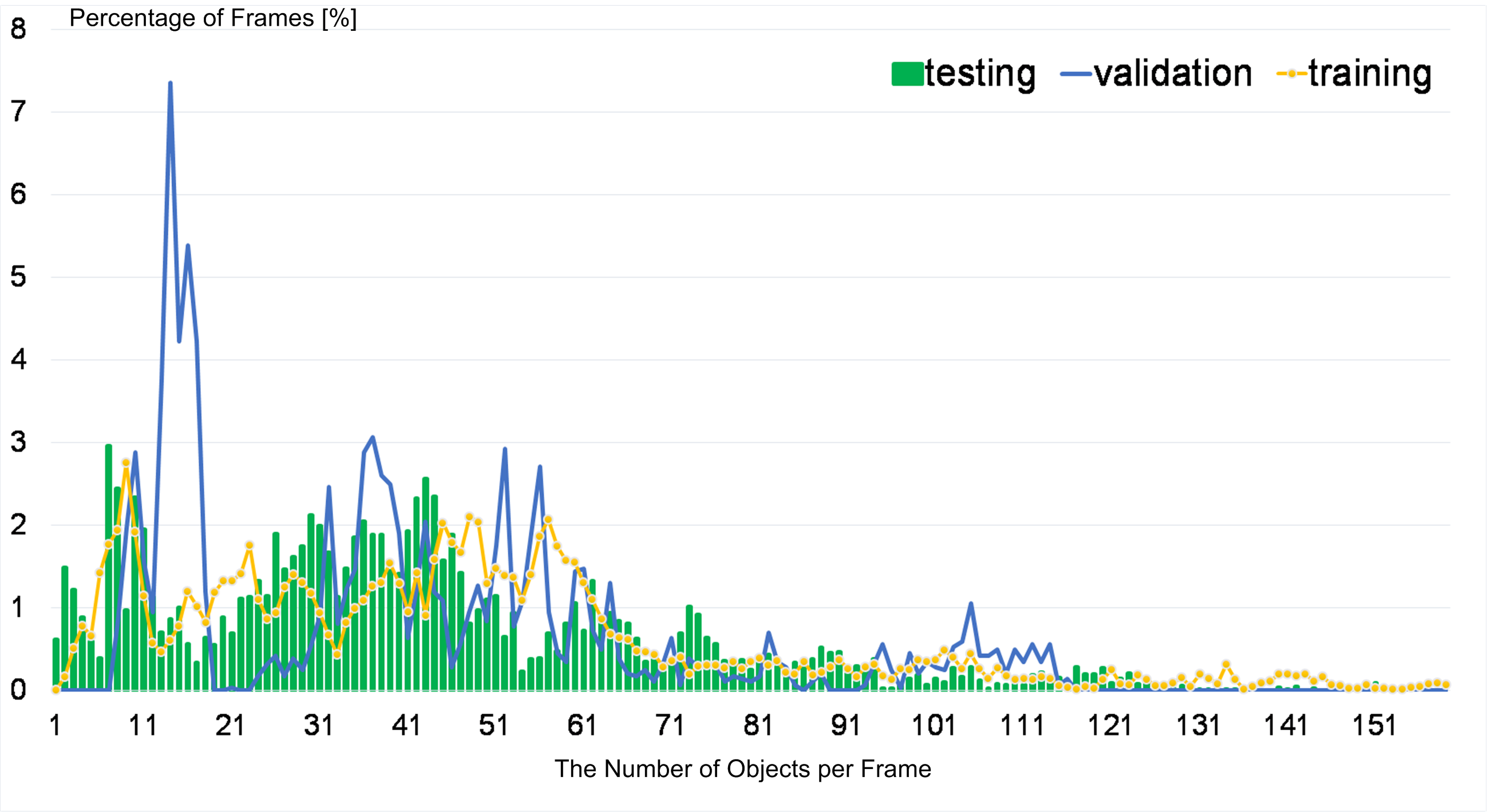}
\vspace{-2mm}
\caption{The number of objects per frame {\em vs.} percentage of frames in the training, validation and testing sets for ({\bf Task 2}) object detection in videos.}
\label{fig:video_number_distribution}
\end{figure*}

Similar to {\bf Task 1}, the task of object detection in videos aims to locate object instances from a predefined set of categories, but the detection is from a \emph{video} instead of a static image as in {\bf Task 1}. Specifically, given a video clip, a detection algorithm is required to produce a set of bounding boxes of each object instance in each video frame (if any), with real-valued confidences. We provide $96$ challenging video clips for the task, including $56$ clips for training ($24,201$ frames in total), $7$ for validation ($2,819$ frames in total) and $33$ for testing ($12,968$ frames in total). The videos of the three subsets are recorded at different locations, but share similar environments and attributes. We plot the number of objects per frame {\em vs.} percentage of frames for training, validation, and testing sets in Figure \ref{fig:video_number_distribution}.

\begin{figure*}[t]
\centering
\includegraphics[width=1.0\linewidth]{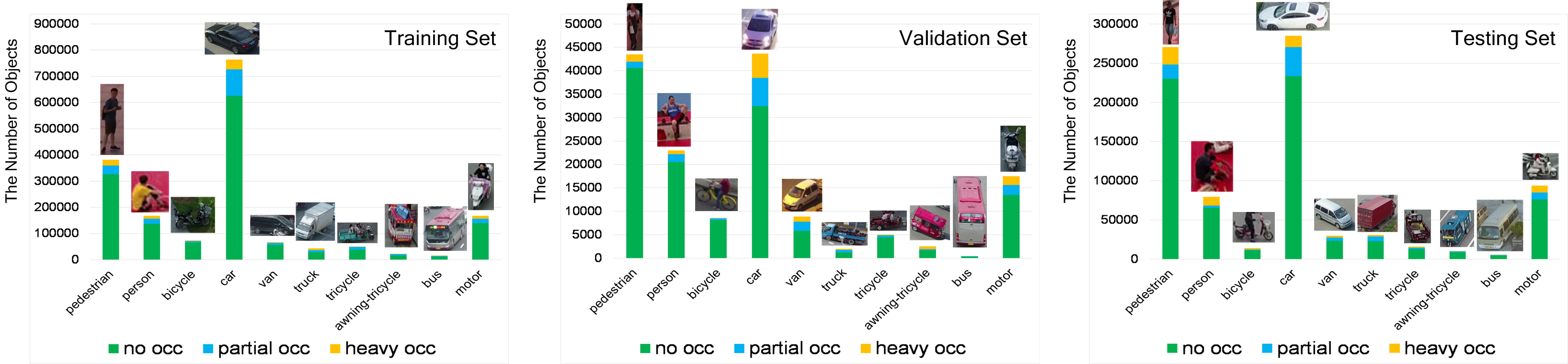}
\caption{The number of objects with different occlusion degrees of different object categories in the training, validation and testing sets for ({\bf Task 2}) object detection in videos.}
\label{fig:video_annotation_statistics}
\end{figure*}

We use the same object categories as that in ({\bf Task 1}) and provide manually annotated ground truth bounding boxes in each video frame. Similar to ({\bf Task 1}), we also provide the annotations of occlusion and truncation ratios of each object. We show some annotated examples in Figure \ref{fig:t2_annotation}, and present the number of objects with different occlusion degrees of different object categories in training, validation, and testing sets in Figure \ref{fig:video_annotation_statistics}.

\subsubsection{Evaluation Criteria}
For {\bf Task 2}, we require each evaluated algorithm to generate a list of bounding box detections with confidences in each video frame. Motivated by the evaluation protocols in MS COCO \cite{DBLP:conf/eccv/LinMBHPRDZ14} and ILSVRC \cite{isvrc-2017}, we use the AP$^{\text{IoU}=0.50:0.05:0.95}$, AP$^{\text{IoU}=0.50}$, AP$^{\text{IoU}=0.75}$, AR$^{\text{max}=1}$, AR$^{\text{max}=10}$, AR$^{\text{max}=100}$ and AR$^{\text{max}=500}$ metrics to evaluate the results of detection algorithms, which is similar to {\bf Task 1}. Notably, the AP$^{\text{IoU}=0.50:0.05:0.95}$ score is used as the primary metric for ranking methods. Please see \cite{DBLP:conf/eccv/LinMBHPRDZ14,isvrc-2017} for more details.

\begin{figure*}[t]
\centering
\includegraphics[width=1.0\linewidth]{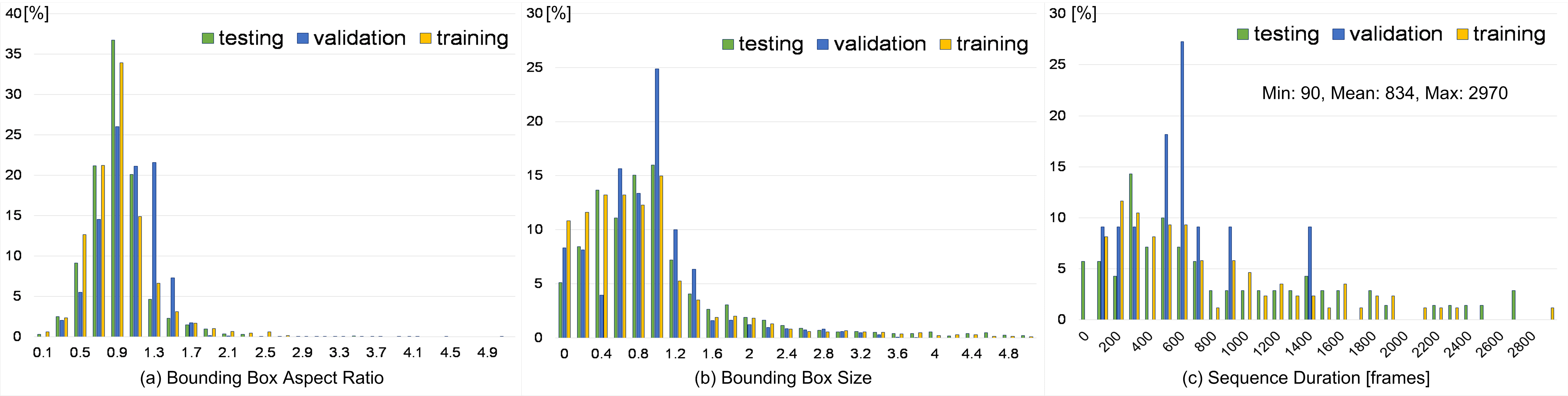}
\vspace{-2mm}
\caption{(a) The number of frames {\em vs.} the aspect ratio (height divided by width) change rate with respect to the first frame, (b) the number of frames {\em vs.} the area change rate with respect to the first frame, and (c) the distributions of the number of frames of video clips, in the training, validation and testing sets for ({\bf Task 3}) single object tracking.}
\label{fig:sot_number_distribution}
\end{figure*}

\subsection{Task 3: Single Object Tracking}
While the term ``object tracking" can be sometimes ambiguous, in {\bf Task 3} we focus on generic single object tracking, also known as model-free tracking. In particular, for an input video sequence and the initial bounding box of the target object in the  first frame, {\bf Task 3} requires a tracking algorithm to locate the target bounding boxes in the subsequent video frames. We provide $167$ video sequences with manually annotated target ground truths. Unlike most previous single object tracking benchmarks, we divide all these video clips into training, validation, and testing sets, with $86$ sequences ($69,941$ frames in total), $11$ sequences ($7,046$ frames in total) and $70$ sequences ($62,289$ frames in total), respectively. The tracking targets in these sequences include pedestrians, cars, buses, and animals. Some annotated examples and the statistics of targets are presented in Figure \ref{fig:t3_annotation} and \ref{fig:sot_number_distribution}.

\begin{figure*}[t]
\centering
\includegraphics[width=0.9\linewidth]{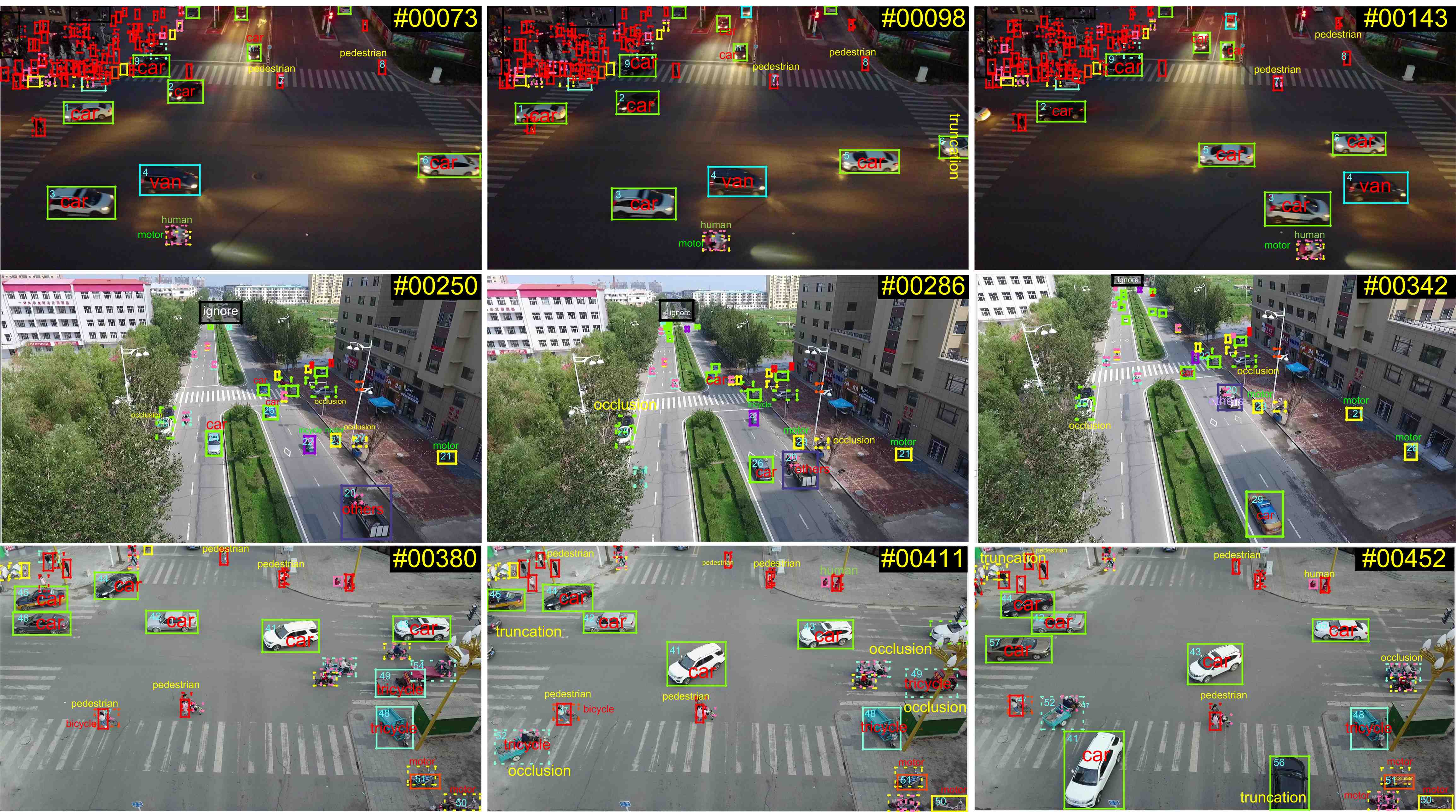}
%\vspace{-2mm}
\caption{Some annotated example video frames of (\textbf{Task 4}) multi-object tracking. The dashed bounding box indicates the object is occluded. Different bounding box colors indicate different classes of objects. The identities of the objects are shown at the left corner of the bounding boxes. For better visualization, we only display some identities of the targets and attributes.}
\label{fig:t4_annotation}
\end{figure*}

\subsubsection{Evaluation Criteria}
For the single object tracking task, the performance is evaluated by the success and precision scores, same as in \cite{DBLP:journals/pami/WuLY15}. Specifically, we plot the percentage of successfully tracked frames {\em vs.} the bounding box overlap threshold, and use the area under the curve (AUC) as the evaluation criterion. Meanwhile, we also plot the percentage curve of frames where the centers of the tracked object are within the given threshold distance to the ground truth, and use the percentage at the threshold of $20$ pixels as the precision score in evaluation. Notably, the success score is used as the primary metric for ranking methods.

\subsection{Task 4: Multi-Object Tracking}
Given an input video sequence, multi-object tracking aims to recover the trajectories of objects in the video. The task uses the same data as in {\bf Task 2} (\ie, object detection in videos). Depends on the availability of prior object detection results, {\bf Task 4} is divided into two sub-tasks, denoted by {\bf Task 4A} (without prior detection) and {\bf Task 4B} (with prior detection). Specifically, for {\bf Task 4A}, an evaluated algorithm is required to recover the trajectories of objects in video sequences without taking the object detection results as input. By contrast, for {\bf Task 4B}, prior object detection results are provided and an evaluated algorithm can work on top of the prior detection. The number of objects {\em vs.} percentage of frames in training, validation, and testing sets are plotted in Figure \ref{fig:video_number_distribution}; some annotated examples are given in Figure \ref{fig:t4_annotation}; and the number of objects with different occlusion degrees of different object categories in training, validation, and testing sets are presented in Figure \ref{fig:video_annotation_statistics}.

\subsubsection{Evaluation Criteria}
For \textbf{Task 4A}, we use the protocol in \cite{isvrc-2017} to evaluate the tracking performance. Specifically, each algorithm is required to output a list of bounding boxes with confidence scores and the corresponding identities. We sort the tracklets (formed by the bounding box detections with the same identity) according to the average confidence of their bounding box detections. A tracklet is considered correct if the intersection over union (IoU) overlap with ground truth tracklet is larger than a threshold. Similar to \cite{isvrc-2017}, we use three thresholds in evaluation, \ie, $0.25$, $0.50$, and $0.75$. The performance of an algorithm is evaluated by averaging the mean average precision (mAP) per object class over different thresholds. Please refer to \cite{isvrc-2017} for more details.

For \textbf{Task 4B}, we use the protocol in \cite{mot-2016} to evaluate the algorithm performance. More specifically, the average rank of $10$ metrics (\ie, MOTA, IDF1, FAF, MT, ML, FP, FN, IDS, FM, and Hz) is used to compare different algorithms. The MOTA metric combines three error sources: FP, FN and IDS. The IDF1 metric indicates the ratio of correctly identified detections over the average number of ground truth and computed detections. The FAF metric indicates the average number of false alarms per frame. The FP metric describes the total number of tracker outputs which are the false alarms, and FN is the total number of targets missed by any tracked trajectories in each frame. The IDS metric describes the total number of times that the matched identity of a tracked trajectory changes, while FM is the total number of times that trajectories are disconnected. Both the IDS and FM metrics reflect the accuracy of tracked trajectories. The ML and MT metrics measure the percentage of tracked trajectories less than $20\%$ and more than $80\%$ of the time span based on the ground truth respectively. The Hz metric indicates the processing speed of the algorithm.

\section{Conclusion}
We introduce a new large-scale benchmark, {\bf \VIS}, to facilitate the research of object detection and tracking on the drone platform. With over $6,000$ worker hours, a vast collection of object instances are gathered, annotated, and organized to drive the advancement of object detection and tracking algorithms. We place emphasis on capturing images and video clips in real life environments. Notably, the dataset is recorded over $14$ different cites in China with various drone platforms, featuring a diverse real-world scenarios. We provide a rich set of annotations including more than $2.5$ million annotated object instances along with several important attributes. The {\bf \VIS} benchmark is made available to the research community through the project website: \url{www.aiskyeye.com}. We expect the benchmark to largely boost the research and development in visual analysis on drone platforms.

%\section*{Acknowledgment}

\bibliographystyle{IEEEtran}
\bibliography{reference}

\ifCLASSOPTIONcaptionsoff
  \newpage
\fi

\end{document}